\pdfoutput=1
\documentclass[letterpaper, 10pt, conference]{ieeeconf}  
\usepackage{multicol}
\usepackage[bookmarks=true]{hyperref}
\usepackage{float}
\usepackage{graphicx}
\usepackage{epstopdf}
\usepackage{booktabs}
\usepackage{tabularx}
\usepackage{multirow}
\usepackage{caption}
\usepackage{subcaption}
\usepackage{amssymb}
\usepackage{color}
\usepackage{amsmath}
\usepackage{amsmath,bm}
\usepackage{comment}
\usepackage{multirow}
\usepackage{comment}
\usepackage{balance}
\usepackage{soul}
\usepackage[ruled, linesnumbered]{algorithm2e} 
\usepackage[bottom]{footmisc}
\usepackage{tabularx}
\IEEEoverridecommandlockouts                              
\usepackage{xcolor}

\newcommand{\mypara}[1]{\vspace{0.7em}\noindent\textbf{#1}}

\usepackage[ruled,linesnumbered]{algorithm2e}

\title{\LARGE \bf
EdgeVO: An Efficient and Accurate Edge-based Visual Odometry
}

\author{ Hui Zhao$^{1,2}$,  Jianga Shang$^1$, Kai Liu$^2$, Chao Chen$^2$, Fuqiang Gu*$^2$  \\
$^1$School of Computer Science, China University of Geoscience, Wuhan, China\\
$^2$College of Computer Science, Chongqing University, Chongqing, China\\
{\tt\small \{zhaohui, jgshang\}@cug.edu.cn,  \{liukai0807, cschenchao, gufq\}@cqu.edu.cn}
\thanks{* Corresponding Author}
}

\begin{document}
\maketitle
\begin{abstract}
Visual odometry is important for plenty of applications such as autonomous vehicles, and robot navigation. It is challenging to conduct visual odometry in textureless scenes or environments with sudden illumination changes where popular feature-based methods or direct methods cannot work well. To address this challenge, some edge-based methods have been proposed, but they usually struggle between the efficiency and accuracy. In this work, we propose a novel visual odometry approach called \textit{EdgeVO}, which is accurate, efficient, and robust. By efficiently selecting a small set of edges with certain strategies, we significantly improve the computational efficiency without sacrificing the accuracy. Compared to existing edge-based method, our method can significantly reduce the computational complexity while maintaining similar accuracy or even achieving better accuracy. This is attributed to that our method removes useless or noisy edges. Experimental results on the TUM datasets indicate that EdgeVO significantly outperforms other methods in terms of efficiency, accuracy and robustness.
\end{abstract}
\section{Introduction}
Visual Odometry (VO), which estimates the camera motion from a stream of images, plays an important role in many applications, such as augmented reality (AR), autonomous vehicles, and robot navigation \cite{cadena2016past,taketomi2017visual,younes2017keyframe}. Popular VO methods include salient feature-based approaches \cite{klein2007parallel,mur2017orb} and direct methods \cite{kerl2013robust,engel2015large,engel2017direct}. Feature-based methods obtain the motion estimation by extracting and matching feature points, while direct methods track the camera based on photometric or geometric error \cite{newcombe2011kinectfusion,pomerleau2011tracking,henry2012rgb}. Feature-based methods are advantageous of being insensitive to brightness change, but are heavily dependent on the textures and computationally expensive. By contrast, direct methods do not require salient texture features and are more computation-friendly, but they are more sensitive to brightness change and vulnerable to local optimum. 
\begin{figure}
	\centering
	\begin{subfigure}[b]{0.45 \linewidth}
		\centering
		\includegraphics[width=1.5 in]{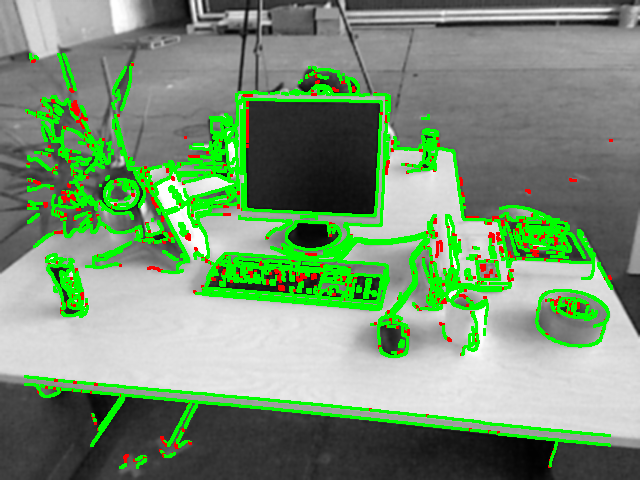} 
		\caption{Features by RESLAM}
	\end{subfigure}
	\quad
	\begin{subfigure}[b]{0.45 \linewidth}
		\centering
		\includegraphics[width=1.5 in]{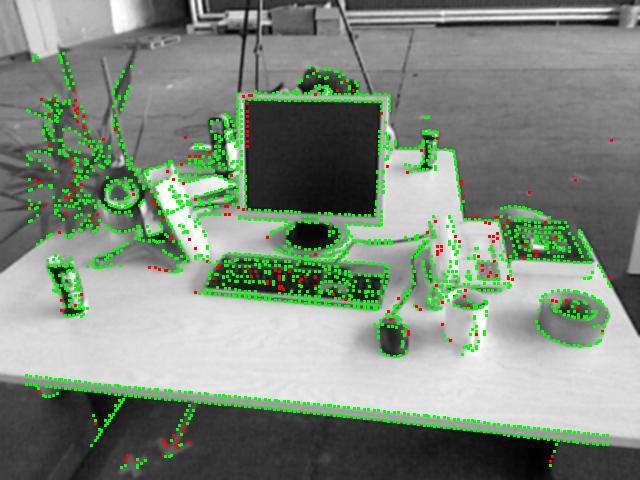} 
		\caption{Features by EdgeVO}
	\end{subfigure}
	
	\vspace{0.08cm}
	\begin{subfigure}[b]{0.45 \linewidth}
	\centering
	\includegraphics[height = 1.2in, width=1.6 in]{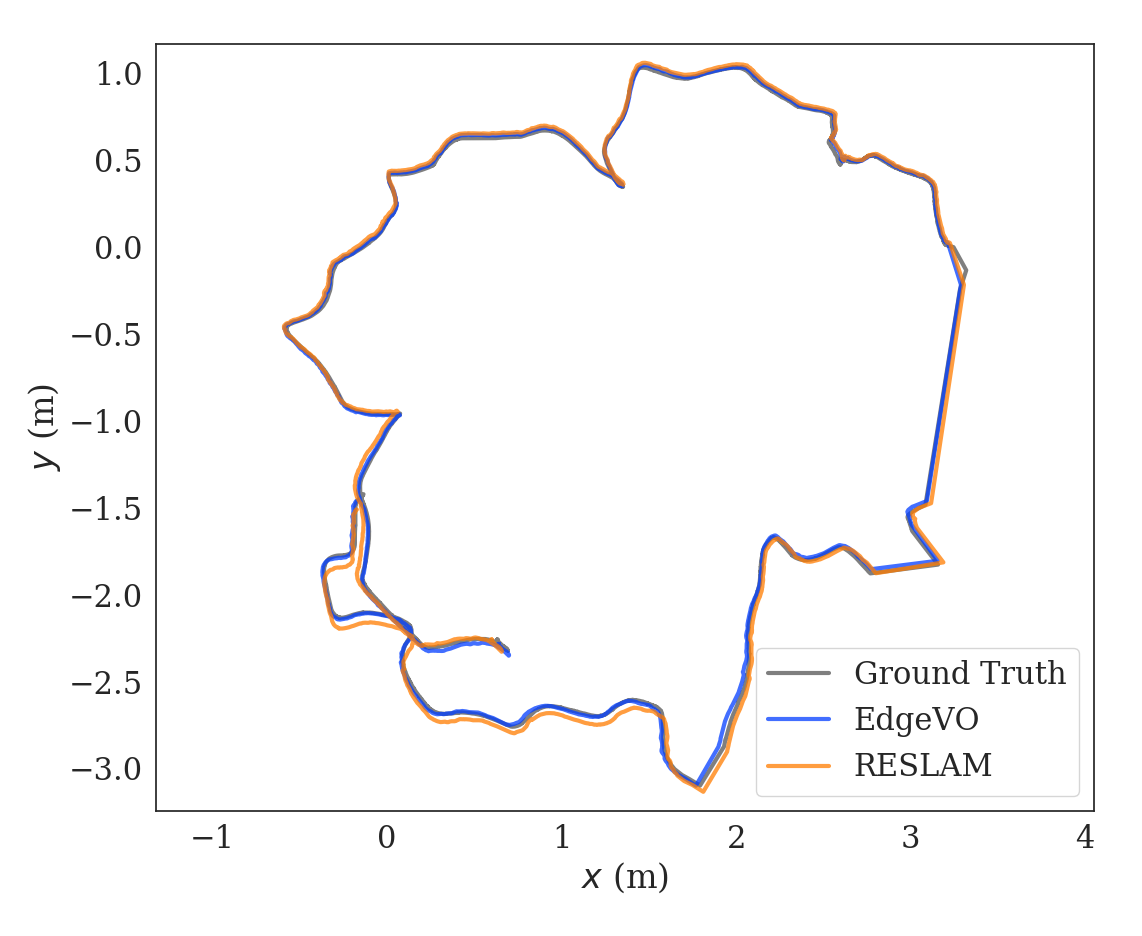}
	\caption{Trajectory comparison}
    \end{subfigure}
    \quad
	\begin{subfigure}[b]{0.45 \linewidth}
	\centering
	\includegraphics[width=1.6 in]{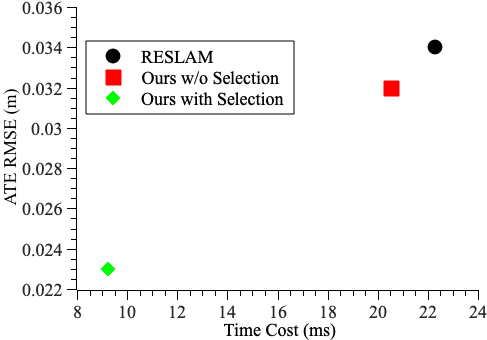}
	\caption{Accuracy vs time}
    \end{subfigure}
	\caption{An illustration of EdgeVO (our method) as compared to RESLAM \cite{schenk2019reslam}. (a) Edge Features extracted by RESLAM. (b) Edge features extracted by EdgeVO (our method). (c) Estimated trajectory by the proposed EdgeVO and RESLAM on the \textit{fr2\_desk} sequence from TUM RGBD datasets. (d) The time cost vs accuracy of the proposed EdgeVO with and without selection as compared to RESLAM on the selected sequence. }
	\label{fig:EdgeVOintroduction} 
\end{figure}

Although many efforts have been made to improve the performance of visual odometry, it is challenging to estimate the camera motion in some corner cases \cite{chen2018comparison,zunjie2018real}. For example, textureless scenes cannot provide sufficient feature points for salient feature-based approaches, and sudden illumination changes can cause the assumption invalidation of photometric consistency for direct methods. These factors reduce the effectiveness of the VO algorithm or even cause tracking failure \cite{kim2018edge,kim2020edge}. To overcome these drawbacks, several solutions have been proposed. Some combine various features, such as lines \cite{zhou2015structslam,pumarola2017pl} and planes \cite{salas2014dense,kaess2015simultaneous,ma2016cpa}, to track the pose of the camera. Another solution is to improve direct methods by using affine lighting correction \cite{engel2015large}, photometric calibration \cite{engel2017direct}, and descriptor fields \cite{quenzel2020beyond} to handle the illumination changes. Multi-metric fusion approaches  \cite{whelan2015elasticfusion,dai2017bundlefusion} combine geometric and photometric to improve the accuracy and robustness of camera tracking to a certain extent.

As another attempt, the recently emerging edge-based methods \cite{schenk2019reslam, kim2020edge, kuse2016robust, zhou2018canny} have shown robust performance in these corner cases. They employ image edges that are observed more naturally and stably in textureless environments, and apply 3D-2D edge registration to estimate the pose of a camera. Compared to feature-based and direct methods, edge-based methods can work in both textureless scenarios and environments with brightness changes. However, there are still some challenges remained in edge-based VO approaches. It is known that edge alignment is less accurate due to the absence of dedicated descriptors for edges, thus producing much higher outlier rates \cite{zhou2018canny}. To improve the accuracy of pose estimation, most edge-based systems exploit a large amount of measurements to solve a nonlinear least-squares problem \cite{tarrio2015realtime,tarrio2019se,schenk2017robust}. In each iteration of pose optimization, the nearest neighbor searching is performed multiple times, which drastically increase the computational complexity of such method.
 
In this paper, we propose a novel method, which is called EdgeVO, for the pose estimation of the camera. Unlike existing edge-based approaches, which use all extracted edges, we use only a small subset of the extracted edges that are selected according to their contribution to motion estimation. Then, we track only the selected edges to estimate camera motion. More specially,
 we formulate edge selection as a submatrix selection problem, and then use an efficient greedy algorithm to approximate optimal results with considering spatial correlation and observational uncertainty. Our approach is proven to achieve $\mathcal{O}(\textit{n})$ computational complexity with 1/2 approximation guarantee. Figure \ref{fig:EdgeVOintroduction} gives an illustration of the proposed EdgeVO with RESLAM \cite{schenk2019reslam}.
 
Compared to existing edge-based methods, our method can significantly reduce the computational complexity while maintaining similar accuracy or even achieving better accuracy since our method removes useless or noisy edges. Figure~\ref{fig:EdgeVOintroduction} compares the number of extracted features, estimated trajectory, and accuracy vs time of EdgeVO and popular RESLAM. We have also conducted experiments on the TUM datasets, experimental results on the TUM datasets indicate that EdgeVO significantly outperforms other methods in terms of efficiency, accuracy and robustness. 

\section{Problem Formulation}
\label{sec:edgealignment}

\label{motion estimation}
For each incoming frame ${I}_c$, we use the iterative closest points (ICP)-based algorithm to estimate the camera motion $\bm{\xi}_{c,r}$ relative to its reference frame ${I}_r$ by aligning their respective edges pixel. The edge alignment is performed by re-projecting valid edge pixel set from ${I}_r$ to ${I}_c$ and minimizing the distance to the closest edge pixels in ${I}_c$. To speed up edge alignment, we first compute the distance field $D_c:\Omega\subset\mathbb{R}^2 \longmapsto \mathbb{R}^+$ of ${I}_c$ using Distance Transform~\cite{felzenszwalb2012distance}. For an edge pixel $\mathbf{p}_i$ with inverse depth ${\rho}_i$ in reference frame, the re-projected distance residual ${r}_i$ under the transformation $\bm{\xi}_{c,r}$ is defined as:
\begin{equation}
r_i=D_c(\omega(\rho_i,\mathbf{p}_i,\bm{\xi}_{c,r})),
\end{equation}
where $\omega(\xi,\mathbf{p},\rho)=\pi(\mathbf{R} \cdot \pi^{-1}(\mathbf{p},\rho)+\mathbf{t})$ is a warping function for re-projecting the edge from ${I}_r$ to ${I}_c$, and $\pi$ is the perspective projection function.

We arrange all residuals of valid edges in the reference frame into one column vector. The residual vector can be formulated as $\mathbf{r}=[r_1 \quad  \cdots \quad r_n]^T$.  The optimal relative camera motion $\bm{\xi}{_{c,r}^*}$ is estimated by minimizing the sum of squared residuals:
\begin{equation}
\label{eqmotionestimation}
\bm{\xi}{_{c,r}^*}=\mathop{\arg\min}_{\bm{\xi}_{c,r}}\\ \mathbf{r}^{\rm T}{\mathbf{W}}\mathbf{r},
\end{equation}
where $\mathbf{W}$ is a weighting matrix. We use a Huber weighting scheme to reduce the influence of large residuals.

Iterative methods such as Gauss-Newton or Levenberg-Marquardt are often used to solve the nonlinear least squares problem. In each iteration, they perform a first-order linearization of $\mathbf{r}$ about the current value of $\bm{\xi}_{c,r}$ by computing the motion Jacobian $\bm{J}$, and then solve the linear least-squares problem to update the motion, namely
\begin{equation}
\mathbf{r}({\xi}_{c,r}) \approx \mathbf{r}({\xi}{_{c,r}^k})+\bm{J}\Delta\xi.
\end{equation}

At the final iteration, the least-squares covariance of the motion estimation is calculated as $\Sigma = (\bm{J}^T\bm{J})^{-1}$, which reveals the uncertainty of motion estimation. Generally, it depends on the spectral properties of $\bm{J}$, if we use more valid edges to track, the singular values of $\bm{J}$ would increase in magnitude, and the accuracy of motion estimation is more likely to be improved~\cite{zhao2018good}. 

However, the number of edges detected in each frame are very large (e.g., tens of thousands), and using all of them would greatly reduce computational efficiency. In this paper, we try to use only a small subset of the edges to speed up motion estimation while preserving the accuracy and robustness. As suggested in~\cite{zhao2018good}, we aim to find a submatrix (i.e., a subset of row blocks) in $\bm{J}$ that preserves the overall spectral properties of $\bm{J}$ as much as possible.

Now, we formulate the edge selection problem. Let $U$ be the indices of row blocks in full matrix $\bm{J}$ and $U=\{0,1,\cdots,n-1\}$. $S$ denotes the index subsets of selected row blocks, $[\bm{J}(S)]$ is the corresponding concatenated submatrix, and $k$ is the number of selected row blocks. Then, the submatrix selection problem is formulated as:
\begin{equation} 
\label{basicselection}
\mathop{\arg\max}_{S \subseteq U}\text{logdet}([\bm{J}(S)]^T[\bm{J}(S)]) \quad \text{subject to}\quad |S|=k,
\end{equation}
where \textit{logdet}($\cdot$) is a submodular function to compute the log determinant of a matrix, which quantifies the spectral properties of the matrix.

It is known that the submodular optimization problem is NP-hard, the stochastic greedy method~\cite{zhao2018good,mirzasoleiman2015lazier,zhao2020good} is commonly used to provide a near-optimal solution with an $(1-1/e-\epsilon)$ approximation guarantee. It starts with an empty set, and in each round $i$, it randomly samples a subset $\mathcal{R} \in U \backslash S$. Then, it picks up an element $e \in \mathcal{R}$ with maximizing the marginal gain:
\begin{equation}
\rho_e(S) = \text{logdet}(S \cup e) - \text{logdet}(S).
\end{equation}


\section{Proposed Method: EdgeVO}
\label{framework}
\begin{figure}
	\centering
    \includegraphics[width= 0.95\linewidth]{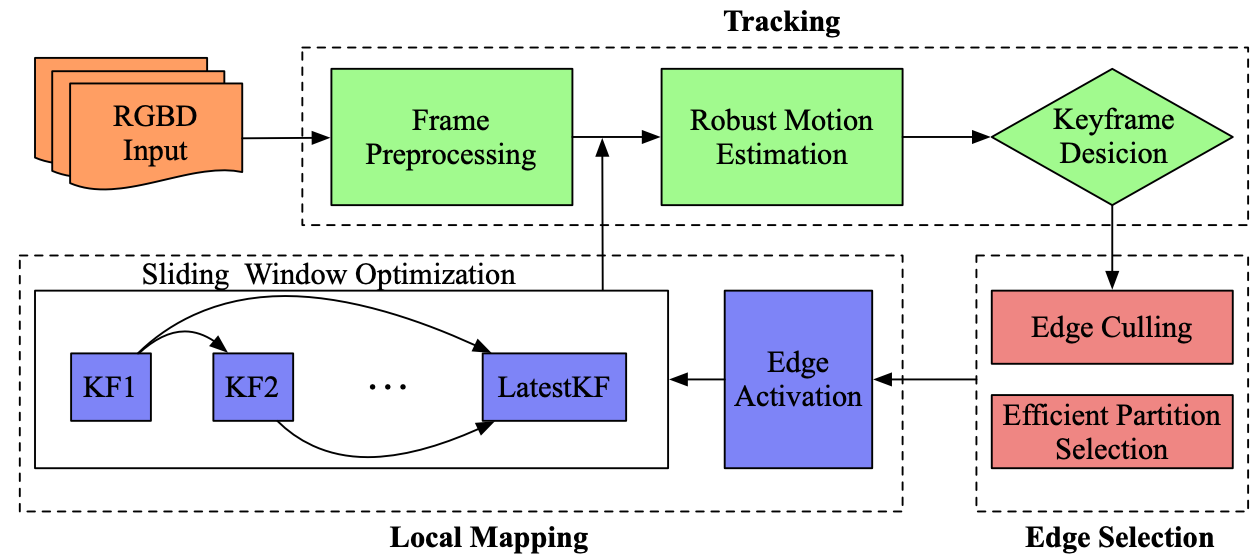}
	\caption{Overview of the EdgeVO architecture. } 
	\label{fig:architecture} 
\end{figure}

 The architecture of EdgeVO is demonstrated in Figure \ref{fig:architecture}. It consists of two parallel threads: tracking, and local mapping. The tracking thread estimates the relative camera motion between the current frame and the latest keyframe and then decides if a new keyframe should be created. If a new keyframe is created, we propagate it to the local mapping thread and refine the relevant state variables using the sliding window optimization. The edges are selected for each keyframe, which are used for both relative motion estimation and local optimization. Next, we elaborate on each key component. 

\subsection{Tracking}
In order to estimate the camera relative motion, we firstly preprocess each frame to get the edges image and distance field pyramid. Then, a coarse-to-fine robust optimization is performed for edge alignment based on the distance field pyramid. Finally, we decide whether a keyframe is created according to the alignment status. In the following, we will provide more details of these steps.

\begin{figure}
	\centering
	\begin{subfigure}[b]{0.3 \linewidth}
		\centering
		\includegraphics[width=1 in]{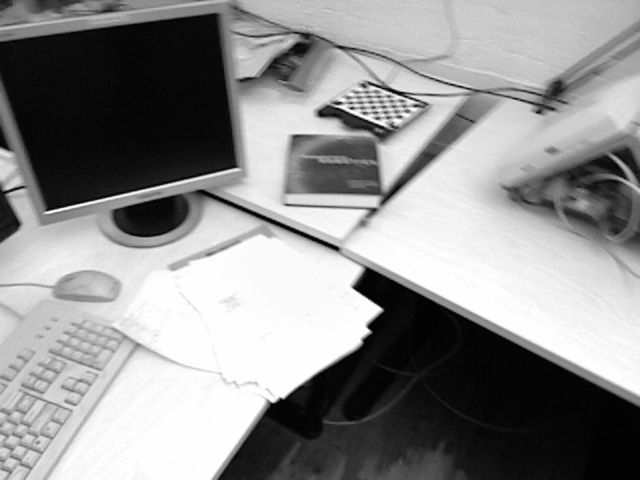} 
		\caption{Gray image}
	\end{subfigure}
	\begin{subfigure}[b]{0.3 \linewidth}
		\centering
		\includegraphics[width=1 in]{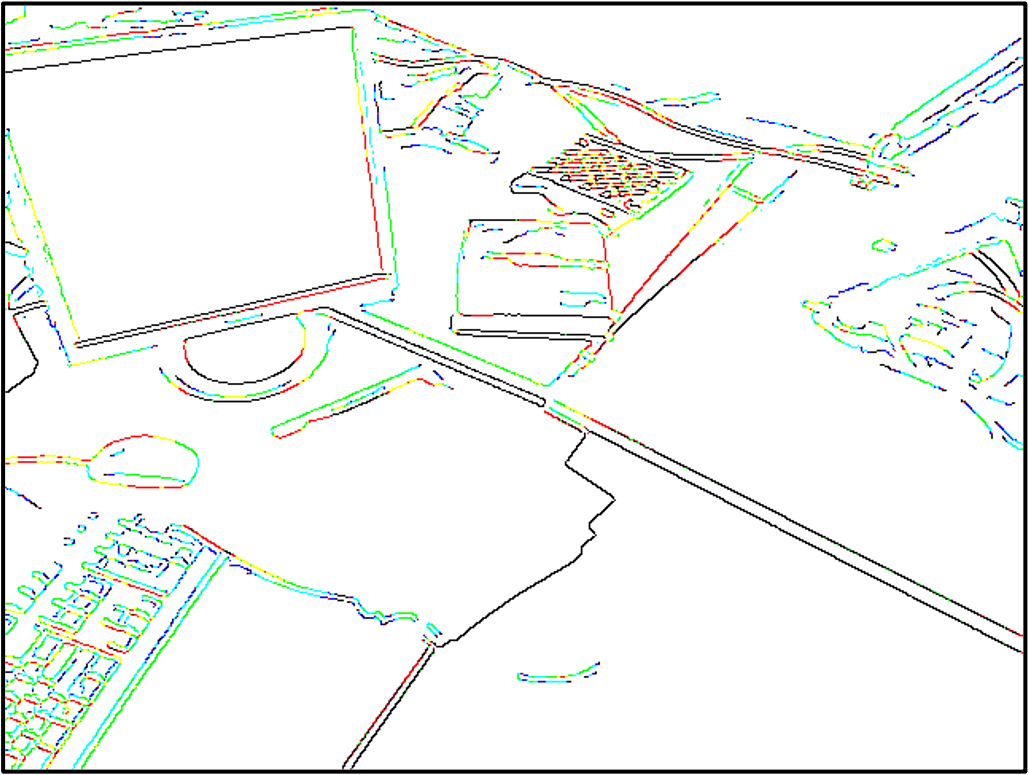} 
		\caption{Edge image}
	\end{subfigure}
	\begin{subfigure}[b]{0.3 \linewidth}
	\centering
	\includegraphics[width=1 in]{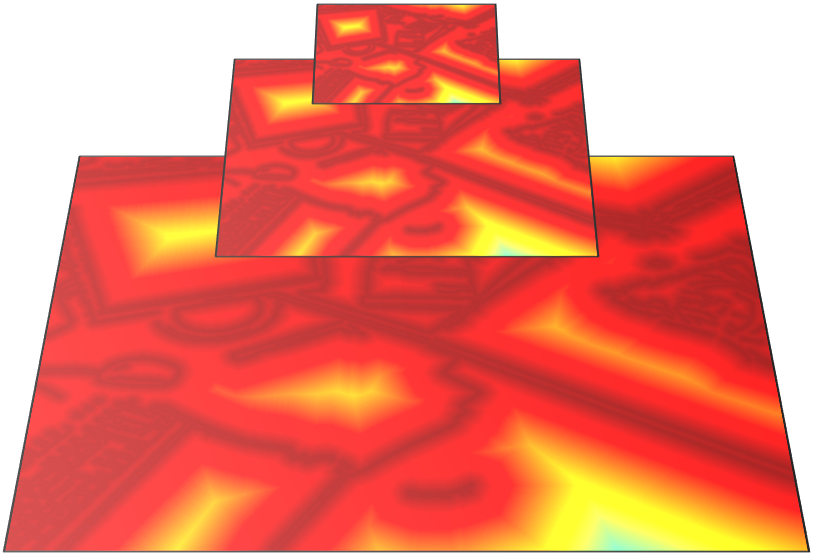} 
	\caption{Distance fields}
    \end{subfigure}
	\caption{The processing flow for each incoming frame. (a) The gray image extracted from the RGBD frame. (b) The corresponding edge image, where its color encodes gradient magnitude information: red-high and blue-low. (c) The distance field pyramid where the color code is red-near and yellow-far.}
	\label{fig:rgb_edge_dt} 
\end{figure}

\mypara{Frame Preprocessing} As shown in Figure~\ref{fig:rgb_edge_dt}, when the current frame comes, we first detect edge pixels using the Canny algorithm~\cite{canny1986computational}. It works well in low-texture and illumination changes scenes because it locally finds the strongest edges by non-maximum suppression of high gradient regions. Then, we compute the distance field of the edge image, and create a three-level distance field pyramid for achieving robust optimization of edge alignment. Instead of alternately performing edge detection and distance transform, we directly generate low-resolution distance field by downscaling from the highest resolution level by linear interpolation~\cite{schenk2019reslam}.

\mypara{Robust Motion Estimation} The relative camera motion $\mathbf{\xi}_{c,k}$ between the current frame ${I}_c$ and the latest keyframe ${I}_k$ is estimated within the ICP framework. Specifically, we use only the selected edges in ${I}_k$, and adopt a coarse-to-fine optimization scheme with the distance field pyramid to handle the large displacement. The edge re-projection and alignment are layer-by-layer performed from the lowest resolution level to the highest level. After final iteration, the optimal camera relative motion $\bm{\xi}{_{c,k}^*}$ is obtained.

In order to improve the robustness and accuracy of camera tracking, we try to identify and remove potential outliers in each iterative optimization. The specific process is as follows. First, the residual terms that exceed a certain threshold are removed from the objective function described in equation ~\eqref{eqmotionestimation}.
We set the threshold separately for each level in the distance field pyramid. Second, if a pair of putative edge correspondences is reasonable, their gradient directions should be as consistent as possible under the assumption that there is no large rotation between ${I}_k$ and ${I}_c$. Note that the closest edge coordinate is additionally retained in extracting the distance field. We reproject an edge pixel $\mathbf{p}$ from ${I}_k$ to ${I}_c$ and query its closest edge $\mathbf{p}'$. Then, the matched pair is regarded as outliers if the inner product of their gradient directions is under a certain margin:
\begin{equation}
\label{gradientfilter}
    \mathbf{g}(\mathbf{p}) \cdot \mathbf{g}(\mathbf{p}')<\eta,
\end{equation}
where $\mathbf{g}(\cdot)$ is the normalized image gradient on the edge pixel. A larger threshold $\eta$ means stricter requirements for consistency, which is empirically set to 0.6 in this work.

\mypara{Keyframe Decision} In general, well-distributed keyframes are very important to ensure the performance of the whole system. Similar to \cite{engel2017direct}, we use the following criteria to decide whether the current frame will be selected as a keyframe: (i) We use the mean squared error of optical flow $t$ from the last frame to the latest frame to measure the changes in the field of view and the mean flow without rotation $t^{'}$ to measure the occlusions. The current frame is selected as keyframe, if $w_1 t + w_2 t^{'}>1$, where $w_1$, $w_2$ provide a relative weighting of two indicators. (ii) If the number of edge correspondences between $I_c$ and $I_k$ is below thirty percent of the average edge correspondences, we treat the current frame as a new keyframe. (iii) If none of the previous conditions occur, we insert new keyframe in a fixed interval (1 second).

\subsection{Edge Selection}
\label{edgeselection}
To increase the efficiency of EdgeVO, a small subset of edges is selected from each new created keyframe, and used for tracking subsequent frames and local mapping. We first cull the useless and spurious edges, and then solve the submatrix selection problem to guide edge selection. Specially, we propose an efficient partition selection approach to make the selection more efficiently and robustly. 

\mypara{Edge Culling} The edges detected by the Canny algorithm are usually redundant. We thus cull a part of the edges to reduce the size of the ground set before edge selection. First, the edges without depth information are culled since they cannot be directly used for re-projection. Then, we filter out the edges whose gradient magnitude is lower than the high threshold in the Canny algorithm. Only the edges with higher gradient will be retained since they have high signal-to-noise ratio, which is more likely to be beneficial for more accurate motion estimation.

\begin{figure}
	\centering
	\begin{subfigure}[b]{0.45 \linewidth}
		\centering
		\includegraphics[height =1.2 in]{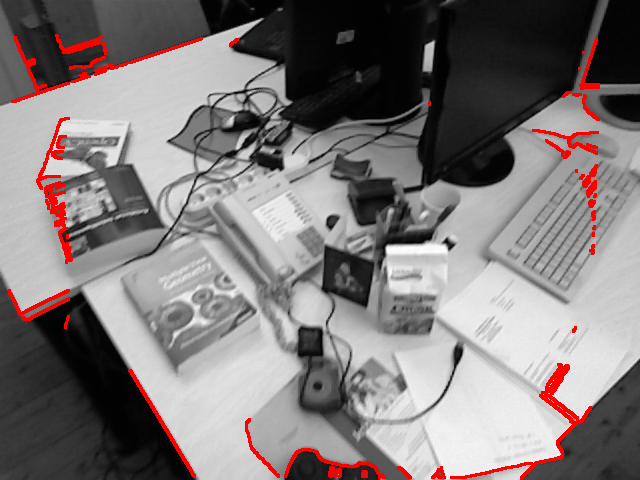} 
		\caption{Clustered edge samples}
	\end{subfigure}
~
	\begin{subfigure}[b]{0.45 \linewidth}
		\centering
		\includegraphics[height=1.2 in]{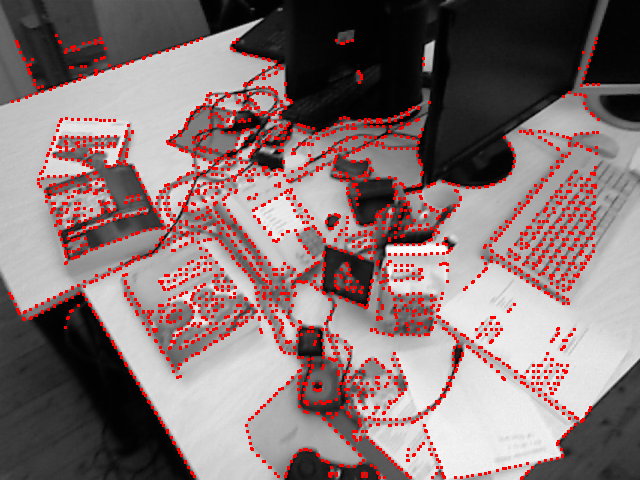}
		\caption{Uniform edge samples}
	\end{subfigure}
	\caption{Comparison of edge selection. (a) Solving the submatrix selection problem~\eqref{basicselection} leads to selected edges cluster in image. (b) If optimize over the partition matriod, the selected edges is well distributed.}
	\label{fig:sampling_comparison}
\end{figure}

\mypara{Efficient Partition Selection} To avoid the clustering of selected edges, we optimize the submatrix selection problem over a partition matroid. Specifically, the indice set $U$ of row blocks in full matrix is divided into $k$ disjoint partitions, where $U = \bigcup_{i=1}^{k} {P}_i$, and the edges with the constraint $|{S} \cap {P}_i|=1$ are selected. We denote $\mathcal{F}$ as a set of all feasible solution ${S} \in \mathcal{F}$, $\mathcal{M}=({U},\mathcal{F})$ as the partition matroid. The edge selection problem is reformulated as:
\begin{equation} \label{partitionselection}
\mathop{\arg\max}_{S \in \mathcal{F}} \text{logdet} (\bm{H}(S))\quad
\text{subject to} \quad |S \cap P_i|=1,
\end{equation}
where $\bm{H}(S) = \bm{J}^T(S)\bm{J}(S)$ is the Hessian matrix of the edges. We impose an imaginary grid on the image and count the edges in each grid to build partitions. This forces the selected edges to spread evenly over the whole image as seen in Figure~\ref{fig:sampling_comparison}.


In order to reduce the computational complexity of submatrix selection, we design a special approximation algorithm, namely Stochastic Partition Greedy. At the ${i}$-th iteration, we randomly pick up a partition $P_i$ from $U$. Then, an edge is selected from the partition when the following condition holds:
\begin{equation}
e^* := \mathop{\arg\max}_{e \in {P}_i} \text{logdet}(\bm{H}(S_{i-1})+\bm{H}(e)+\lambda\bm{I}),
\end{equation}
where ${S}_{i-1}=\{e_1,\cdots,e_{i-1}\}$ is the subset selected at previous iteration. The introduction of diagonal matrix $\lambda\mathbf{I}$ is to improve its numerical stability. In this way, each edge is computed only once, and the complexity is thus reduced to $\mathcal{O}(n)$. 

\textbf{Proposition 1} \textit{We use $T$ to denote the optimal solution for the problem \eqref{partitionselection}, and use ${S}$ to denote the approximation solution provided by Stochastic Partition Greedy. The approximation guarantee is thus  $f(S) \geq {1/2}f(T)$.}

In order to avoid invalid selection, we first assume that if the edge is re-observed in the current frame, it is valid to constrain the motion estimation. We model the case by introducing a Bernoulli distribution ${B}=\{b_0,\cdots,b_{n-1}\}$, where ${b_i}$ is a binary variable. ${b_i}=1$ means that the $i$-th edge is re-observed in the current frame, while ${b_i}=0$ means that such edge is absent in the current frame. Then we re-write the submodular function as,
\begin{equation}
f(S) = \sum\nolimits_{i=1}^k b_i \rho_{e_i}(S_{i-1}).
\end{equation}

It is clear that if the edge $e_i$ is re-observed, then $b_i=1$ and it is able to provide valid marginal gain. Otherwise, this marginal gain simply disappears and we try to find other valid edges to maximize the function. Since ${b_i}$ is a random variable, the value of the function \textit{logdet} is a stochastic quantity. Hence, we should maximize the expectation of this function. Let ${\mathbb{E}(b_i)}={p_i}$ denote the probability of the edge $e_i$ being re-observed, and we have that,
\begin{equation}
\mathbb{E}(f(S)) = \sum\nolimits_{i=1}^k p_i \rho_{e_i}(S_{i-1}).
\end{equation}

We can judge whether an edge is re-observed or not from two aspects. First, as the camera moves, some edges extend beyond the camera's field of view, and it would not be re-observed for the subsequent frames. In the practice, we perform the visibility check and filter out these invisible edges based on the prior camera motion. Then, the more distinctive the appearance of the edge, the more likely it is to be re-observed. Thus, we model the probability of the edge $e_i$ being re-observed as,
\begin{equation}
p_i = \frac{1}{1+e^{a-m_i}}, 
\end{equation}
where $a$ is the high threshold in the Canny algorithm, and $m_i$ is the gradient magnitude of the edge $e_i$. 

In this way, the gradient magnitude of an edge is larger than the high threshold. It is thus more likely to be detected by the Canny algorithm in the next frames, which is a desired behavior. Finally, we provide the pseudo-code of the complete edge selection approach in Algorithm \ref{alg:EPS}.
\begin{algorithm}
    \caption{Efficient Partition Selection}
    \label{alg:EPS}
    \LinesNumbered
    \KwIn{${P}$=$\{P_1,\cdots,P_{k}\}$, the prior motion $\xi$;}
    \KwOut{the edge set $S$}
    Initialize the set $S=\varnothing$\;
    \While{$P \neq \varnothing$}{
    Randomly choose a partition $P_i$ from $P$\;
    \ForEach{$e_j \in P_i$}{
    \eIf{the edge is visible}{Compute the probability $p_j$ and $\bm{H}(j)$;}{\textbf{Continue;}}
    $e^* = \mathop{\arg\max}\limits_{e \in {P}_i}~{p_j}\cdot \text{logdet}(\bm{H}(S_{i-1})+\bm{H}(e)+\lambda\mathbf{I})$\;
    $S_i \leftarrow{S_{i-1} \cup {e^*}}$\;
    $P \leftarrow{P\backslash P_i}$\;
    $\bm{H}(S_i) = \bm{H}(S_{i-1})+\bm{H}(e^*)$\;}}
    \textbf{Return}\quad$S$.\
\end{algorithm}
\subsection{Local Mapping}
To improve the consistency of the camera trajectory, we maintain a small local window $\mathcal{W}$ of keyframes in the local mapping component, where $W$ is set to a value between 5 and 7. For each new keyframe, the edges that satisfy certain criterions in the window are activated to create new geometric constraints, and then we perform a sliding window optimization to jointly refine the inverse depths of all active edges, global camera poses and the intrinsic $\mathbf{c}$ within the window. The specific steps of local mappings are as follows.

\mypara{Edge Activation} When a new keyframe is added, we use it to activate the edges of the previous keyframes in the window. In order to obtain the evenly distributed edges and reliable geometric constraints, we divide the image in cells into fixed size (e.g., $20 \times 20$ pixels). We then reproject all edges into these grids, while the active edges are selected from each cell. The following criterions need to be satisfied at the same time to activate an edge:
\begin{enumerate}
    \item The candidate edge’s geometric residual cannot exceed a certain threshold, which is set to the median of all residuals;
    \item The reprojected gradient direction of the candidate edge should be as consistent as possible with the image gradient direction of its closest edge in the new frame. We set the angle between the two directions to be in the range $[0^\circ, 30^\circ]$;
    \item The edges that are tracked for longer period of time are considered to be more reliable. Therefore, we count the number of times each edge was successfully tracked and select the older one as the active edge;
\end{enumerate}

\mypara{Silding Window Optimization} For the keyframe $I_i$, $I_j$ in the window, the geometric residuals are computed by reprojecting the active edges in $I_i$ into $I_j$,
\begin{equation}
    r=D(p^{'}(\rho,\mathbf{\xi}_i,\mathbf{\xi}_j,\mathbf{c})),
\end{equation}
where $p^{'}$ is the projected position of $p \in S$ in the keyframe $I_j$, and $\mathbf{\xi}_i$, $\mathbf{\xi}_j$ are the estimated camera poses relative to the world frame. All state variables in the window are denoted by $\boldsymbol{\chi}$. The optimal state vector is estimated through minimizing the overall residuals over the window:
\begin{equation}
    \boldsymbol{\chi}^{*}=\mathop{\arg\min}\limits_{\boldsymbol{\chi}} \sum\limits_{i\in \mathcal{W}}\sum\limits_{\mathbf{p}\in S}\sum\limits_{j\in \mathcal{W}\backslash i} w(r)r^2.
\end{equation}

We fix the bound size of the local window. One of previous keyframes need to be marginalized before adding a new one. Following \cite{engel2017direct}, we keep the latest two keyframes in the window, and marginalize a keyframe if it is further away from the newest one or it has less edges visible in others. Before marginalizing one keyframe, we first adapt the marginalization strategy with the Schur complement to marginalize its active edges and the edges that are not observed in the last two keyframes. This is to retain the sparsity of our method. 

\section{Expriments and Results}
\label{experiment}
\subsection{Experimental Setup}
In this section, we evaluate the performance of our method on the standard RGBD benchmark dataset--TUM RGBD \cite{sturm2012benchmark}, which is widely used for evaluating various RGBD VO and SLAM algorithms. The dataset provides synchronized RGB images and depth images recorded with a Microsoft Kinect v.$1$ sensor at $30$Hz. The ground truth trajectories are obtained by a motion capture system. All experiments are implemented on a desktop computer with an Intel CPU of i$5-8400$ CPU and a RAM of $16$GB.
\begin{table}
	\centering
	\caption{\label{tab:accuracy} \centering Comparison of the Absolute Trajectory Error on TUM-RGBD dataset}
	\resizebox{.5\textwidth}{!}{
	\begin{tabular}{cccccc}
		\toprule[0.5pt]
	    Sequence & ElasticFusion & ORB-SLAM2 & Canny-VO & RESLAM & Ours \\ 
		\midrule[0.5pt]
		fr1\_desk &0.020	&\textbf{0.016}	&0.044	&0.036	&0.018  \\ 
		fr1\_desk2 &0.048	&\textbf{0.030}	&0.187	&0.063	&0.046 \\ 
		fr1\_floor &$\times$  &$\times$	&\textbf{0.021}	&0.040	&0.033 \\
		fr1\_rpy  &\textbf{0.025}	&0.031	&0.047	&0.048	&0.037 \\
		fr1\_xyz  &0.011	&0.020	&0.043	&0.024	&\textbf{0.009} \\
		fr2\_desk	&0.071	&0.027	&0.037	&0.034	&\textbf{0.023} \\
		fr2\_rpy	&0.015	&\textbf{0.003}	&0.007	&0.006	&0.006 \\
		fr2\_xyz	&0.011	&0.005	&0.008	&0.005	&\textbf{0.003} \\
		fr3\_office	&\textbf{0.017}	&0.021	&0.085	&0.078	&0.033\\
		fr3\_nostr\_tex\_near	&\textbf{0.016}	&0.030	&0.090	&0.199	&0.026\\
		fr3\_str\_notex\_far	&0.030	&$\times$	&0.031	&0.084	&\textbf{0.024}\\
		fr3\_str\_tex\_far	&0.013	&0.012	&0.013	&0.015	&\textbf{0.011}\\
		fr3\_str\_tex\_near	&0.015	&0.014	&0.025	&0.013	&\textbf{0.012}\\
		fr3\_cabinet	&$\times$	&$\times$	&0.057 	&0.058 	&\textbf{0.024}\\
		\bottomrule[0.5pt]
	\end{tabular}
	}
\label{tab:acc_36}
\end{table}

\subsection{Experimental Results}
We compare the proposed EdgeVO with several baseline methods in terms of accuracy, robustness and efficiency. These baseline methods range from direct methods, feature-based methods, to edge-based methods. The specific baseline methods selected are as follows:
\begin{enumerate}
    \item \textbf{ElasticFusion}~\cite{whelan2015elasticfusion} is a map-centric SLAM approach, which combines geometric and photometric error for frame-to-model pose estimation.
    \item \textbf{ORB-SLAM2}~\cite{mur2017orb} is a feature-based SLAM approach, which estimates camera motion based on feature matches.
    \item \textbf{Canny-VO}~\cite{zhou2018canny} is an edge-based visual odometry method. It adapts the approximate nearest neighbor fields or oriented nearest neighbor fields to estimate camera relative motion. 
    \item \textbf{RESLAM}~\cite{schenk2019reslam} is the first open-source edge-based SLAM system, where the camera motion is estimated based on Euclidean distance fields.
\end{enumerate}

 Note that the odometry and mapping are separated in the methods of ORB-SLAM2 and RESLAM. We deactivate the loop modules for fair comparison, and the results of Canny-VO \cite{zhou2018canny} and ElasticFusion \cite{whelan2015elasticfusion} are directly taken from the original papers. We run ORB-SLAM2 and RESLAM on computing platform and report their performance. 

\mypara{Estimation Error} We first compare the achieved error of EdgeVO with the selected baseline methods. The Root Mean Squared Error (RMSE) of the translational component of the Absolute Trajectory Error (ATE) is taken as the performance metric. 
The results are shown in Table~\ref{tab:accuracy}, from which we can see that the proposed EdgeVO performs the best. It witnesses a lower error than all the baseline methods in most cases. We can also find that EdgeVO performs much better than other edge-based methods, namely Canny-VO and RESLAM. There are two potential reasons why our method outperforms the two edge-based methods. The first reason is that we remove the potential outliers based on appearance similarity, and activate the edges that are more reliable to optimize the camera motion, which are not considered in Canny-VO and RESLAM. The other reason is that we use only a small number of edges that are selected with a certain strategy since too many edges and uneven distribution may lead to unstable optimization values, and the edge selection could further improve the accuracy. EdgeVO also generally outperforms ORB-SLAM2 and ElasticFusion especially in textureless scenarios, while it performs slightly worse than ORB-SLAM2 and ElasticFusion in feature-rich environments.

\begin{figure}
	\centering
    \includegraphics[width= 0.95\linewidth]{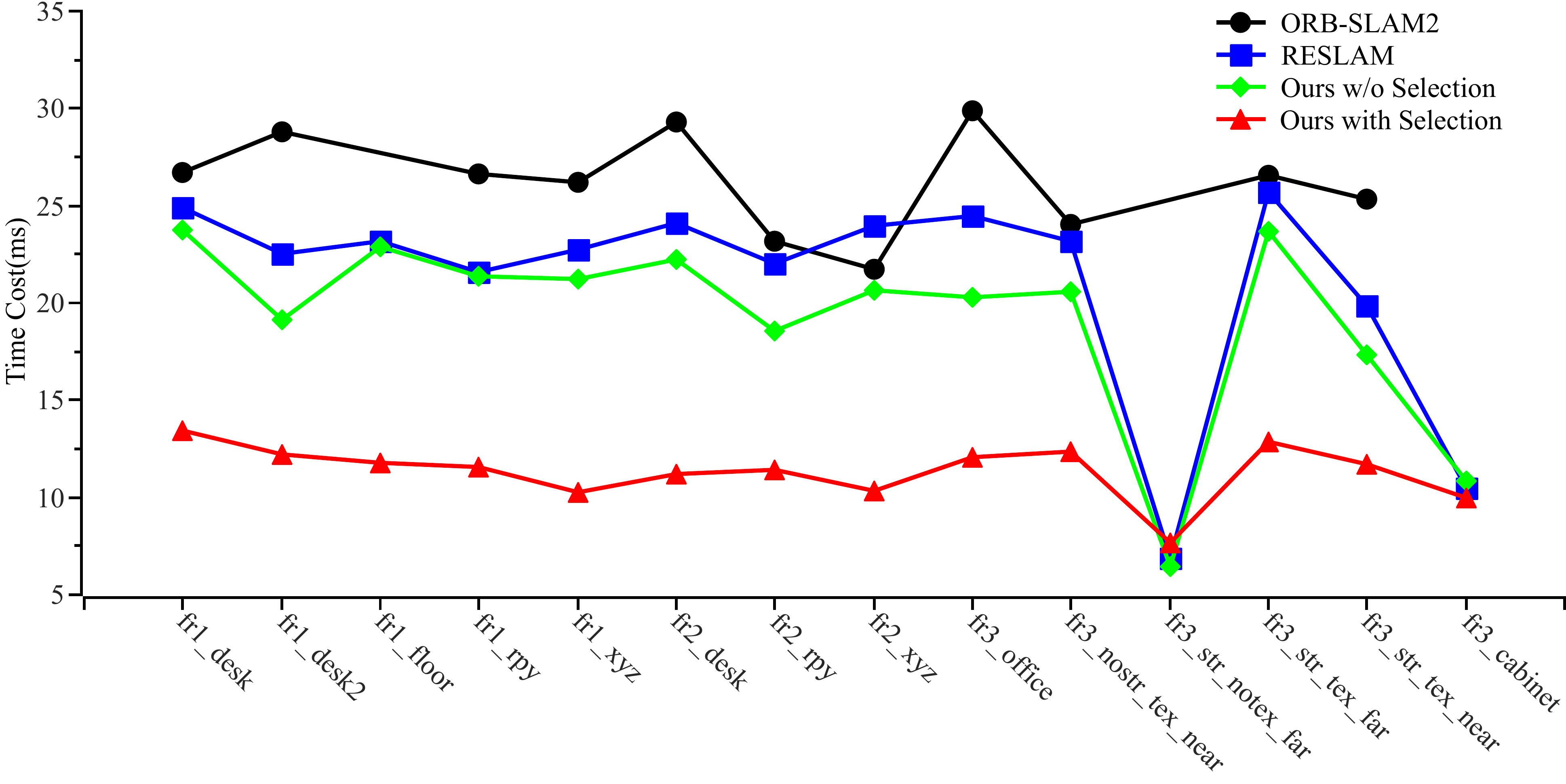}
	\caption{Comparison of computational cost} 
	\label{fig:system_time_cost} 
\end{figure}

\mypara{Efficiency} Then, we compare the computational efficiency of EdgeVO with the baseline methods. We calculate the average running time per frame of ORB-SLAM2, RESLAM and our method on each sequence, the results are shown in Figure~\ref{fig:system_time_cost}. It is clear that the proposed EdgeVO has the lowest time cost and could achieve about $80$Hz reporting the result on the majority of sequences, due mainly to the fact that we use only a small subset of edges in the tracking and local mapping. If our method use all edges the same as RESLAM does, then the time costs of the two method are close. Compared to the edge-based methods including EdgeVO and RESLAM, ORB-SLAM2 requires feature matches based on the similarity of descriptors and additionally frame-to-model pose optimization, so its computational cost is the highest. Compared to ElasticFusion that requires a GPU and achieves a frequency of $20$Hz for motion estimation, our method is about four times faster in terms of the computing speed even using only a i$5-8400$ CPU. Since Canny-VO is not open-sourced, we directly use the frequency of reporting estimation result, which is about $25$Hz even using a i$7-4770$ CPU which is higher in configuration than the CPU we used. Yet, our method is still more efficient than Canny-VO. In addition, the computational cost of our method is close to that of RESLAM in \textit{fr3\_str\_notex\_far} and \textit{fr3\_cabinet}. This is because the number of edges is inherently less, and edge selection has to keep most of the edges to sufficiently constrain the motion estimation.

\mypara{Robustness} Here, we qualitatively evaluate the robustness of the baseline methods based on their tracking failures. From the Table~\ref{tab:accuracy}, we can find that edge-based methods work successfully for each sequence, while ORB-SLAM2 and ElasticFusion fail in some cases. In detail, the tracking of ORB-SLAM2 is lost in the sequences with sudden motion \textit{fr1\_floor} and poor texture (\textit{fr3\_str\_notex\_far, fr3\_cabinet}), since there are not sufficient and reliable feature matches for tracking. For ElasticFusion, sudden motion makes the optimization difficult to converge. Besides, illumination changes (\textit{fr2\_desk}) and high reflection (\textit{fr3\_cabinet}) negatively affect the photometric-based optimization. Thus, ElasticFusion fails sometimes in tracking or results in a low accuracy in these scenarios. While these limitations do not exist on edge-based methods, the edges in these sequences are sufficient even under poor texture scenes. The optimization based on distance field paradigm has a larger radius of convergence and is insensitive to illumination changes, so edge-based methods show better robustness than ElasticFusion and ORB-SLAM2 in these scenarios.

\subsection{Abalation Study}
\begin{figure}
	\centering
    \includegraphics[width= 0.95\linewidth]{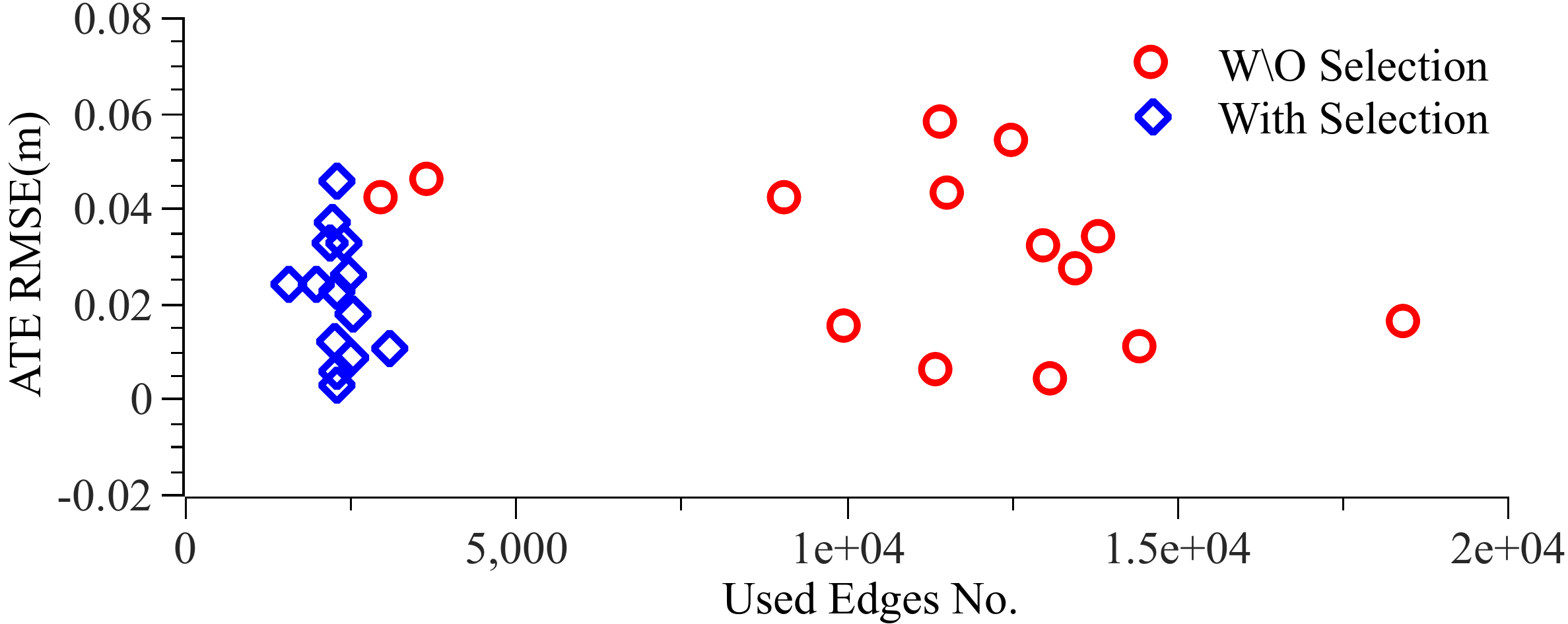}
	\caption{The number of edges used vs the resulted error in each sequence} 
	\label{fig:Nedges_ATE} 
\end{figure}

In this section, we evaluate the effectiveness of the proposed edge selection method by reporting the performance of EdgeVO with and without edge selection. As can be seen from Figure \ref{fig:system_time_cost}, EdgeVO with edge selection has lower running time compared to the one without selection. Additionally, we compare the number of used edges and the resulted RMSE of the translational ATE in each sequence. The results are shown in Figure~\ref{fig:Nedges_ATE}, where each point represents the performance on a sequence. It is clear that our system with edge selection utilizes less edges and achieves similar or higher accuracy than that without selection. 

\section{Conclusion}
In this paper, we propose an accurate, efficient and robust approach for edge-based visual odometry using RGBD cameras, which is called EdgeVO. It can significantly reduce the number of edges required for motion estimation, and result in great computational efficiency improvement over existing edge-based methods without sacrificing accuracy and robustness. We evaluate the proposed method on the public TUM RGBD benchmark, finding that it performs better than state-of-the-art methods with respect to efficiency, accuracy and robustness. For future work, we will investigate large-scale camera motion estimation as well as loop closure for edge-based methods.

\section*{Acknowledgement}
This work was supported by the National Natural Science Foundation of China (No.
42174050, 62172066), Venture \& Innovation Support Program for Chongqing Overseas Returnees (No. cx2021047, cx2021063), and Startup Program for
Chongqing Doctorate Scholars (No. CSTB2022BSXM-JSX005).

\bibliographystyle{IEEEtran}
\balance
\bibliography{manuscript}

\end{document}